\documentclass[letterpaper, 10 pt, conference]{ieeeconf}

\IEEEoverridecommandlockouts                              

\usepackage{graphicx}
\usepackage{bbm}
\usepackage{mathtools}
\usepackage[colorinlistoftodos, textwidth=2cm, textsize=tiny]{todonotes}
\usepackage{float}
\usepackage{amsmath} 
\newcommand{\qedsymbol}{\rule{0.4em}{0.4em}}
\usepackage{algorithm}
\usepackage[noend]{algpseudocode}
\algnewcommand\algorithmicforeach{\textbf{for each}}
\algdef{S}[FOR]{ForEach}[1]{\algorithmicforeach\ #1\ \algorithmicdo}

\usepackage{amssymb}  

\title{\LARGE \bf
Near-optimal irrevocable sample selection for  periodic data streams with applications to marine robotics
}
\author{Genevieve Flaspohler$^{1, 2}$, Nicholas Roy$^{1}$, and Yogesh Girdhar$^{2}$
\thanks{$^{1}$ Computer Science and Artificial Intelligence Laboratory (CSAIL), Massachusetts Institute of Technology, Cambridge MA, 02139, USA {\tt\small \{geflaspo, nickroy\}@mit.edu}}%
\thanks{$^{2}$ Department of Applied Ocean Physics and Engineering, Woods Hole
Oceanographic Institution, Woods Hole MA, 02543, USA {\tt\small yogi@whoi.edu}}%
}

\begin{document}
\maketitle
\thispagestyle{empty}
\pagestyle{empty}
\begin{abstract}
We consider the task of monitoring spatiotemporal phenomena in real-time by deploying limited sampling resources at locations of interest irrevocably and without knowledge of future observations.  This task can be modeled as an instance of the classical \textit{secretary problem}. Although this problem has been studied extensively in theoretical domains, existing algorithms require that data arrive in random order to provide performance guarantees. These algorithms will perform arbitrarily poorly on data streams such as those encountered in robotics and environmental monitoring domains, which tend to have spatiotemporal structure. We focus on the problem of selecting representative samples from phenomena with \textit{periodic} structure and introduce a novel sample selection algorithm that recovers a near-optimal sample set according to any monotone submodular utility function.  We evaluate our algorithm on a seven-year environmental dataset collected at the Martha's Vineyard Coastal Observatory and show that it selects phytoplankton sample locations that are nearly optimal in an information-theoretic sense for predicting phytoplankton concentrations in locations that were not directly sampled. The proposed periodic secretary algorithm can be used with theoretical performance guarantees in many real-time sensing and robotics applications for streaming, irrevocable sample selection from periodic data streams.
\end{abstract}

\section{Introduction}
Many interesting phenomena vary on spatial and temporal scales that are too large to monitor in their entirety. Attempting to understand these phenomena using limited representative samples is known as \textit{constrained sample selection} or \textit{experimental design} \cite{Muller2006}. In most problem formulations, samples are chosen to maximize some utility function while satisfying a fixed cost requirement: an autonomous underwater vehicle (AUV) may need to maximize the amount of phytoplankton in $10$ collected water samples; a planetary rover may need to collect a maximally diverse set of rock samples that weigh less than $5$ \textit{kg}; a policy maker may wish to infer pollutant flow throughout a water body but only spend $\$10,000$ on  water samples. Constrained sample selection problems arise in many real-world contexts, spanning domains from robotics to data mining to online auctions.

\begin{figure}[t]
    \centering
    \includegraphics[width=\linewidth]{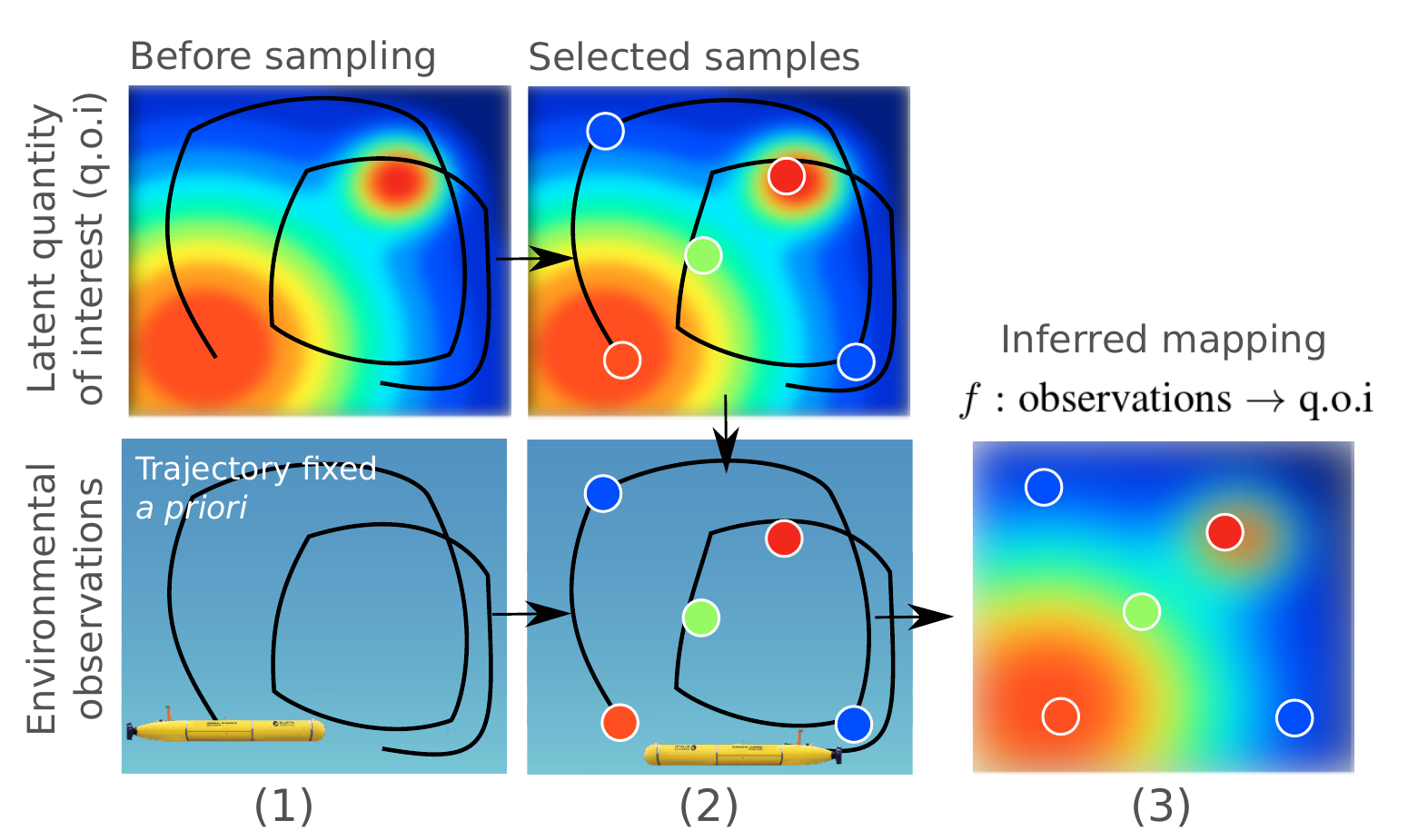}
    \caption{\textbf{Streaming irrevocable sample selection:} In an example of streaming, irrevocable sample selection, an autonomous underwater vehicle must irrevocably collect representative water samples along a fixed trajectory at locations that are the most informative about a latent quantity of interest (q.o.i.), e.g. plankton concentrations (1). After observing the value of the quantity of interest at the sample locations (2), we can infer a mapping between environmental observations and the latent q.o.i. for later use (3).}
\label{fig:title}
\end{figure}

Constrained sample selection problems can be divided into offline and streaming problems. In offline problems, potential sample locations are known ahead of time and an algorithm can make arbitrarily many passes through these locations to find the optimal placement of samples. In streaming problems, potential sample locations are revealed to the algorithm sequentially, and the algorithm must choose to collect or not collect a sample in real-time. Both the offline and streaming constrained sample selection problems are known to be {NP}-hard, but polynomial time approximation schemes exist for a variety of problem formulations \cite{Nemhauser1978}.

One important variant of the streaming sample selection problem arises when an autonomous agent must choose to collect samples \textit{irrevocably}, i.e. the agent must decide in real-time whether to collect a sample, cannot return to collect a sample at a previously rejected location, and cannot later reject a collected sample. This streaming, irrevocable-choice variant of the constrained sample selection problem arises frequently in real-time domains and is known as the secretary problem because of parallels to the problem of hiring the most qualified secretarial candidate from a stream of applicants \cite{Ferguson1989}. Often, in these streaming sampling problems, the quantity of interest (q.o.i) is not directly observable and samples must be selected based on observable quantities that are hypothesized to be correlated with the quantity of interest. In these problems, it is also often desirable to select samples that are informative for the purpose of inference about the distribution of the q.o.i.; this can be addressed by optimizing an \textit{information-theoretic utility function} \cite{Krause2008}. 

For example, an AUV following a fixed trajectory through a marine environment may be equipped with $k$ single-use water samplers and need to collect the set of water samples that are the most informative about the distribution of a quantity of interest (q.o.i.) e.g. plankton species. However, the AUV is unable to measure the plankton present in the water stream in real-time. Instead, the AUV can measure the surrounding environmental conditions and decide to collect a sample based on these environmental covariates (Figure \ref{fig:title}). Finding the optimal set of locations to sample at along its trajectory without a model of the environment is a hard problem: if the AUV collects samples too early, it will not be able to sample the interesting locations it discovers later in the mission; if the AUV passes over interesting locations at the start of the mission, it may not see enough high quality locations later in the mission at which to collect samples.



The secretary problem has a long history and a variety of near-optimal solutions for different problem domains have been developed \cite{Ferguson1989}. However, solutions to the secretary problem nearly always require that data are seen in random order. This stringent requirement is rarely met in robotics and real-time sensing domains, which produce spatially and temporally correlated data streams. In this work, we focus on data streams with periodic spatiotemporal structure. Periodic data arise commonly in environmental monitoring datasets due to natural cycles on a daily, lunar, and annual basis and in robotics tasks such as repetitive surveying. In this work, we introduce a multiple-choice secretary algorithm to choose $k$ samples from a data stream with periodic structure and provide a lower bound on the utility of the selected samples. 




\vspace{0.2em}
The contributions of this work include: \begin{itemize}
    \setlength\itemsep{0.2em}
    \item We introduce the \textit{periodic secretary algorithm}, which leverages spatiotemporal structure to choose near-optimal samples from a periodic data stream according to any monotone submodular utility function.
    \item We develop an algorithmic framework that allows information-theoretic utility functions to be used within secretary algorithms in order to minimize the uncertainty in estimates of a quantity of interest at locations or times that are not directly observed.
    \item We validate our algorithms on a dataset containing phytoplankton observations from January 2009 to January 2016, and show that the phytoplankton samples selected by the periodic secretary algorithm are best able to predict phytoplankton concentrations in environmental conditions that were not directly sampled.
\end{itemize}


\section{Related Work}
The problem of constrained sample selection has been given a thorough treatment in both the offline and streaming settings. In offline settings, previous work has explored using information theoretic utility functions in spatiotemporally correlated data domains to select high utility samples. Nemhauser et al. \cite{Nemhauser1978} show that for submodular utility functions, a simple iterative greedy algorithm where the highest-utility sample given previous samples is selected at each iteration will produce a set with utility greater than $(1 - 1/e)$ times the utility of the optimal set. Other works use this greedy algorithm along with Gaussian process (GP) models and information-theoretic utility functions to do offline sample selection \cite{Krause2008} and to plan information-rich exploration paths for robots \cite{Binney}. There also is a rich body of literature in the spatial statistics community discussing optimal sensor placement in an offline setting for a variety of placement criteria \cite{Muller2006}.


On the other hand, streaming, irrevocable sampling algorithms remain largely constrained to simple utility functions and random arrival order assumptions. When selecting a single maximal sample, the problem is known as the \textit{secretary problem} and  Lindley \cite{Lindley1961} provides a well known result: by observing the maximum utility sample in the first $1/e$ fraction of the stream and picking the first sample with higher utility, the highest utility sample will be selected in $1/e$ cases. If we instead want the set of $k$ samples with maximum utility, the problem is known as the \textit{multiple-choice secretary problem}. Babaioff et al. \cite{Babaioffa} introduce an $e$-competitive algorithm for the multiple-choice secretary problem and an alternative approach \cite{Kleinberg} provides a $1 / (1 - 5 / \sqrt{k})$-competitive algorithm. However, neither algorithm can be implemented with information-theoretic utility functions and both require that data arrive in random order. The most general solution to the multiple-choice secretary problem is presented by Bateni et al. \cite{Bateni2010}. For any monotone submodular utility function Bateni et al. \cite{Bateni2010} provide a $7 / (1 - 1/e)$-competitive algorithm known as the \textit{submodular secretary algorithm}. The submodular secretary algorithm splits the data into $k$ segments and runs the single secretary algorithm on each segment. Despite allowing flexibility in utility function, this algorithm still requires that data arrive in random order.

Kesselheim et al. \cite{Kesselheim} attempt to relax the assumption that data arrive in random order and define a class of distributions for which the assumption is violated but the performance of the standard secretary algorithm remains bounded. However, most spatiotemporally correlated data, including periodic data, do not satisfy even these relaxed constraints. Vardi \cite{Vardi2014} proposes a secretary algorithm for quasi-periodic data which arrive in random order. This algorithm requires that each observation will appear exactly $m$ times in the data stream, an often unrealistic assumption in noisy data streams.

Streaming, irrevocable-choice algorithms have been applied to select samples in environmental monitoring and robotics applications, even when the data streams in question violate random arrival order assumptions. Das et al. \cite{Das} apply the submodular secretary algorithm on-board an AUV to select $k$ water samples with the highest concentration of a harmful phytoplankton and use a GP model to predict these concentrations. However, they directly apply the submodular secretary algorithm, despite their data being spatially correlated, which could lead to arbitrarily poor sampling performance. Girdhar et al. \cite{Girdhar2010} also deploy a modified multiple-choice secretary algorithm on an AUV to choose the most informative images to send back to a ground station. However, this approach is incompatible with the use of information-theoretic utility functions, requires that data arrive in random order, and does not account for spatiotemporal structure in an image stream.

\section{Technical Background}
In the general constrained sample selection problem, we must choose a set $\cal A$ consisting of $k$ sample locations from a finite set of possible locations in our observation space $\cal V$ such that a utility set function $f : 2^\mathcal{V} \to \mathbb{R}^+$ is maximized (Eq. \ref{eq:utility}). The full observation space $\cal V$ is split into a set of locations where it is possible to collect samples $\cal S$ and a set where no samples can be collected $\mathcal{U} = \mathcal{V \setminus S}$. 
\begin{equation}
\label{eq:utility}
    \mathcal A^* = \text{argmax}_{\mathcal{A}\subseteq \mathcal{S}: \left\vert{\mathcal{A}}\right\vert = k} f(\mathcal{A})
\end{equation}
In the offline setting, $\mathcal{S}$ and $\mathcal{U}$ are defined by accessibility, price, or other concerns. In the streaming setting, $\mathcal{S}$ is the set of observations encountered in the data stream. The observation space $\cal V$ can consist of geographic locations or locations in an environmental sensor space, e.g. temperature.

\subsection{Utility functions and tradeoffs}
A variety of utility functions appear in the sample selection literature, including maximizing the sum of utilities of the collected samples \cite{Babaioffa}, maximizing the minimum distance between samples \cite{Girdhar2010,Zhu2016}, maximizing the reduction in entropy ${H(\cdot)}$ over $\mathcal{V}$,  i.e. the \textit{entropy criterion} \cite{Krause2008}: 
\begin{equation}
\label{eq:H}
    f_H(\mathcal{A}) = - H(\mathcal{V} \setminus \mathcal{A} \mid \mathcal{A}) = H(\mathcal{A}) - H(\mathcal{V}),
\end{equation}
or maximizing the mutual information $I(\cdot; \cdot)$ between sampled locations and the rest of the observation space, i.e. the \textit{mutual information criterion}  \cite{Krause2008}:
\begin{equation}
\label{eq:MI}
    f_I(\mathcal{A}) = I(\mathcal{V} \setminus \mathcal{A}; \mathcal{A}) = H(\mathcal{V} \setminus \mathcal{A}) - H(\mathcal{V} \setminus \mathcal{A} \mid \mathcal{A}).
\end{equation}
The entropy and mutual information utility functions directly quantify how useful a sample will be for the task of inference about a quantity of interest that is distributed across the observation space. These information-theoretic utility functions have been widely used to decide optimal placements of sensors in the kriging and spatial statistics literature \cite{Muller2006}. The mutual information criterion seeks to maximize the mutual information between a set of sampled locations $\cal A$ and the rest of the observation space $\cal V \setminus A$. Intuitively, the mutual information criterion reflects how informative the sampled locations are about the rest of the space for the purposes of inference. However, calculating the mutual information criterion requires a model of the entire observation space $\mathcal{V}$ and generally requires $O(\left\vert \mathcal{V} \right\vert^3)$ operations to compute a single time. This can be challenging or impossible to compute in streaming contexts. 

The entropy criterion seeks simply to maximize the reduction in entropy over the observation space by maximizing the entropy of the selected sample set $\mathcal{A}$, since the entropy of the sample space $H(\mathcal{V})$ is constant. The entropy criterion does not depend on knowledge of the entire observation space and can be calculated in $O(k^3)$ operations, where $k$ is maximum cardinality of the selected sample set $\mathcal{A}$. Despite the compelling argument made in favor of the mutual information criterion in \cite{Krause2008}, for real-time applications run on computationally constrained devices, the entropy criterion is an efficient alternative to the mutual information criterion.



\subsection{Submodular set functions}
For an arbitrary utility function $f(\mathcal{A})$, the maximization problem in Eq. (\ref{eq:utility}) is {NP}-hard for both the offline and streaming scenarios \cite{Ko1995}. Fortunately, many commonly used utility functions, including the entropy criterion  \cite{Srinivas2012}, have special structure that allows near-optimal polynomial time approximation schemes. This structure is submodularity \cite{Nemhauser1978}.


\vspace{0.10cm}
\noindent
\textbf{Definition 1} (Submodularity) A set function $f:2^\mathcal{V} \to \mathbb{R}$ is submodular if for every $A \subseteq B \subseteq \cal V$ and $e \in \mathcal{V} \setminus B$, $f(A \cup \{e\}) - f(A) \geq f(B \cup \{e\}) - f(B)$.

Submodularity formalizes the intuitive notion of diminishing returns: the benefit you get from adding a new sample to a large set is less than the benefit you get from adding that new sample to a smaller subset. Monotone submodular utility functions have many beneficial properties: they can be minimized efficiently and near-optimal constrained maximization is possible in polynomial time. We will exploit this structure to develop performance guarantees for our periodic, irrevocable sample selection algorithm using the entropy criterion.

\section{Proposed Model}
\begin{figure}[b]
    \centering
    \includegraphics[width=\linewidth]{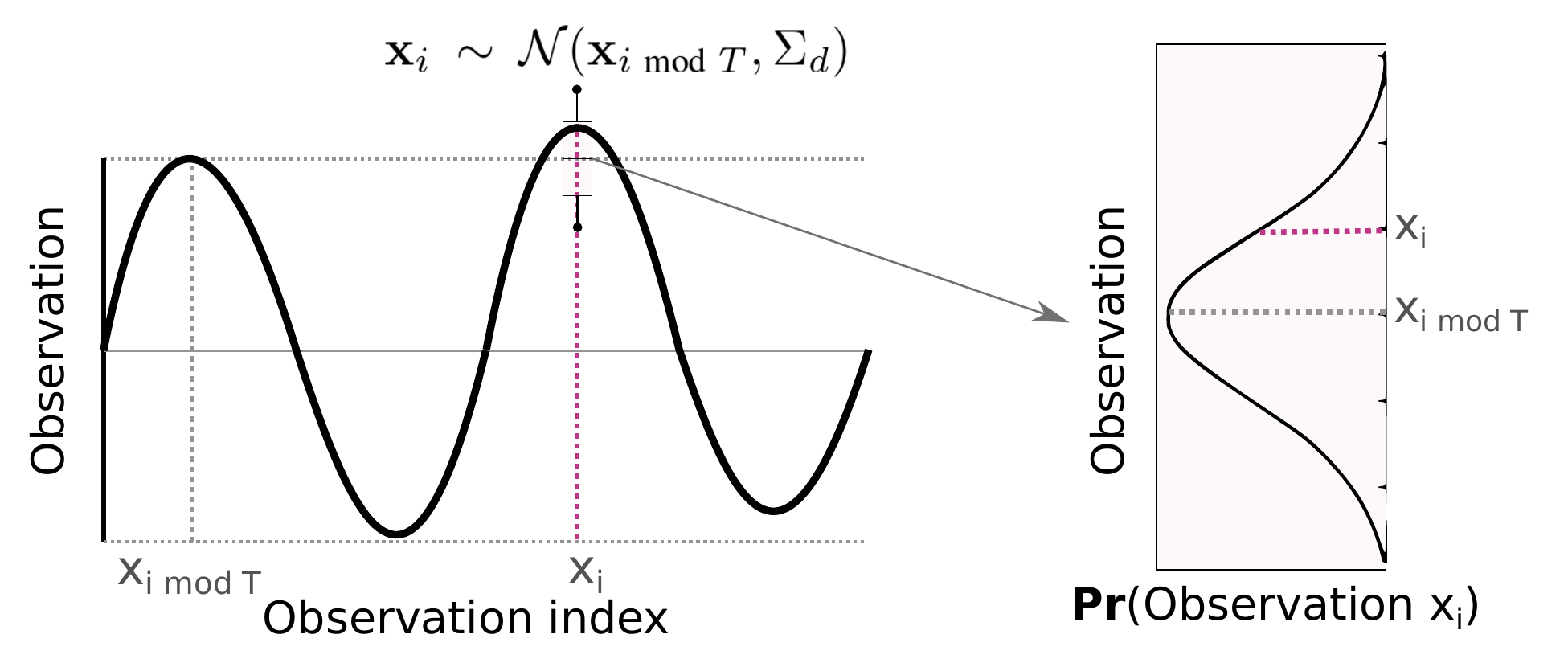}
    \caption{ \textbf{Approximately periodic data:} Our algorithm assumes data are approximately periodic with period $T$ and noise $\Sigma_d$, where the observation $x_i$ at index $i$ is drawn i.i.d. from a Gaussian distribution with mean equal to the observation $x_{i\text{ mod }T}$ and covariance $\Sigma_d$ i.e. $x_i \sim \mathcal{N}(x_{i \text{ mod }T}, \Sigma_d)$}
\label{fig:approx}
\end{figure}

Let the dataset $\mathcal{S}=\{\mathbf{x}_i\} \subseteq \mathcal{V}$ be a stream of observations, such that $\mathbf{x}_i$ is observed at time step $i$. Let $y_i$ be the corresponding latent quantity of interest (q.o.i.) value at time step $i$, which cannot be measured \textit{in vivo} but can be sampled for offline analysis. We define an observation data stream $\mathcal{S}$ to be approximately periodic with period $T$ and noise $\Sigma_d$ if the (possibly vector-valued) observation $\mathbf{x}_i$ at index $i$ is drawn i.i.d. from a Gaussian distribution with mean equal to the observation $\mathbf{x}_{i\text{ mod }T}$ and covariance $\Sigma_d$ i.e. $\mathbf{x}_i \sim \mathcal{N}(\mathbf{x}_{i \text{ mod }T}, \Sigma_d)$ for $i \geq T$ (Figure \ref{fig:approx}). The utility of approximately periodic observations will also be approximately periodic with period $T$ and some utility noise $\sigma_u^2$ for any deterministic utility function $f(\cdot)$. 

Our sampling goal is to select a set of $k$ sample locations $\mathcal{A} \subseteq \mathcal{S}$ that maximally reduce the entropy (Eq. \ref{eq:H}) in predictions of the q.o.i. $y$ over the observation space $\mathcal{V}$. Computing the entropy criterion requires a model of how a latent quantity of interest $y$ is correlated with observations $\mathbf{x}$. Given that the physical sensors in robotics and environmental monitoring domains are noisy, this model would ideally be probabilistic and include a measure of uncertainty in its predictions. Following \cite{Krause2008}, we use a Gaussian process model (GP), a nonparametric generalization of the multivariate Gaussian distribution. A GP model allows us to make predictions about the q.o.i. at new observation locations in the data stream $\mathcal{S}$ based on a set of noisy samples, and to compute the uncertainty in these predictions.

Let $\mathcal{A}_m \subseteq \mathcal{S}_i$ be the set of $m$ observations we have sampled from the first $i$ observations in the data stream. At time step $i + 1$, we must irrevocably decide whether to add the observation to the sample set based on the value of the entropy criterion at that observation. To calculate the entropy criterion at a potential observation $\mathbf{x}_{i+1}$, we must calculate the conditional entropy of that observation, given the locations in observation space of previously collected samples. In a GP model, we can calculate the differential entropy at $\mathbf{x}_{i+1}$ in closed form \cite{Krause2008}:
\begin{equation}
h(\mathbf{x}_{i+1} \mid \mathcal{A}_m) =  \frac{d}{2}\text{ln}(2\pi e) + \frac{1}{2}\text{ln}(\sigma^2_{\mathbf{x}_{i+1}} \mid \mathcal{A}_m)
\label{eq:condH}
\end{equation}
where $d$ is the dimension of $\mathcal{V}$ and $\sigma^2_{\mathbf{x}_{i+1}} \mid \mathcal{A}_m$ is the conditional variance of the GP model at point $\mathbf{x}_{i+1}$ \cite{Rasmussen2006}. Crucially, $\sigma^2_{\mathbf{x}_{i+1}} \mid \mathcal{A}_m$ depends solely on the covariance function used in the GP model and the locations of the samples $\mathcal{A}_m$ in observation space, not on the sampled quantity of interest values at these locations. This important property of GPs allows us to do streaming entropy calculations even if we are unable to observe the value $y_{i+1}$ of a sample at observation $\mathbf{x}_{i+1}$ until post-processing. For our model, we use a squared exponential (SE) covariance function with maximum likelihood parameters estimated from a previous dataset. Although our data are periodic, we do not use a periodic covariance function \cite{Marchant2014}. The periodic covariance function is used in data domains where the latent q.o.i. is periodic but the observations themselves are not; the SE covariance function is sufficient for our model because both the environmental observations and the latent q.o.i are assumed to be periodic with the same period.


After a sampling mission is completed and samples have been collected at various locations in observation space, we use the resulting dataset $\mathcal{D} = \{(\mathbf{x}_i, y_i) \mid \mathbf{x}_i \in \mathcal{S}\}$, $\left\vert \mathcal{D} \right\vert = k$, consisting of observations $\mathbf{x}_i$ and noisy quantity of interest samples at those observations $y_i$, to predict the distribution of the latent quantity of interest at unsampled locations in the observation space using the formulas for the conditional predictive mean and variance of a GP \cite{Rasmussen:2006:GP}. 

\section{Proposed Sampling Algorithm}
Periodic phenomena occur ubiquitously in biological domains due to natural cycles on a daily, monthly, and annual basis and in repetitive robotics tasks. Existing streaming, irrevocable-choice sampling algorithms will perform arbitrarily poorly on these data streams due to their non-random spatiotemporal structure. Given a GP model that allows us to compute the entropy utility function for observations in our data stream, we propose a novel variant of the multiple-choice secretary algorithm for data with approximately periodic structure.

\begin{algorithm}[b]
\caption{Periodic secretary algorithm}
\label{alg:alg1}
\textbf{Input:} Utility function $f$, data stream $\mathcal{S}=\{\mathbf{x}_i\}$, sampling capacity $k$, data period $T$, parameter $\lambda \in \mathbb{R}^+$ \\
\textbf{Output:} Sample set $\mathcal{A} \subseteq \mathcal{S}$ \
\begin{algorithmic}[1]
\Procedure{periodic secretary algorithm}{}
\State $\mathcal{A} \gets \emptyset$
\State $U_R \gets \{f(\{\mathbf{x}_i\}), \text{ for } i \in [0,T)$
\State $\text{threshold} \gets \text{max}(U_R) - \lambda$
\ForEach{$i \in [T,\dots,N]$} 
    \If {$f(\{\mathbf{x}_i\} \cup \mathcal{A}) \geq \text{threshold}$}
    \State $\mathcal{A} \gets \mathcal{A} \cup \mathbf{x}_i$
    \If {$\left\vert  \mathcal{A} \right\vert  = k$} \textbf{return} $\mathcal{A}$
    \EndIf
    \State $U_R \gets \{f(\{\mathbf{x}_i\} \cup \mathcal{A})\}, \text{ for } i \in [0,T)$
    \State $\text{threshold} \gets \text{max}(U_R) - \lambda$
    \EndIf
\EndFor
\State \Return $\mathcal{A}$
\EndProcedure
\end{algorithmic}
\end{algorithm}



Assuming that the period $T$ of an approximately periodic data stream is known or can be estimated, the proposed periodic secretary algorithm consists of two stages. During the initial observation period, the first $T$ observations in the data stream are saved into a reference set ${U}_R$ but no samples are collected. Then, for the remainder of the data stream, our goal is to iteratively sample the next observation in the stream with the highest utility given previously selected samples.  This goal is difficult to achieve without knowledge of the future observations in the data stream. However, for approximately periodic data streams, our algorithm can exploit the information it gathers during the initial observation period to make informed decisions about when to sample.



\begin{figure}[t]
    \centering
    \includegraphics[width=8.0cm]{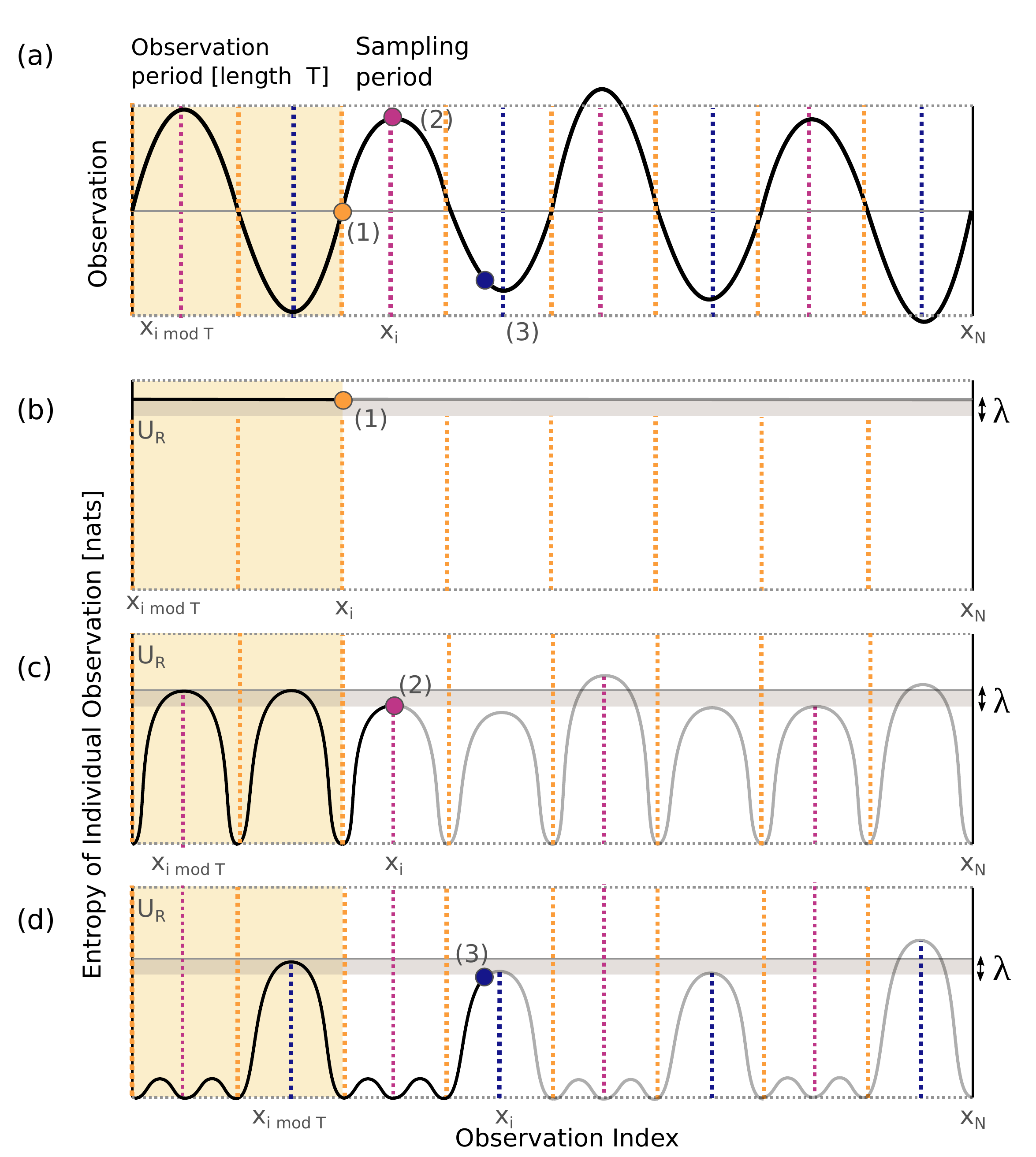}
    \caption{\textbf{Periodic secretary algorithm with threshold parameter $\lambda$:} \textbf{(a)} An approximately periodic data stream with known period T. The first three samples selected using the periodic secretary algorithm are shown. \textbf{(b)} Observed points are in black; unknown future observations are in grey. When the algorithm begins and before any samples have been selected, every subsequent observation has equal entropy [utility], hence the algorithm chooses the first observation after the reference set $U_R$ as the first sample. \textbf{(c)} Given the first sample (1), the utility function is approximately periodic. The algorithm then samples the first observation with entropy [utility] $\geq$ the maximum entropy observation in the reference set $U_R$ minus $\lambda$ (2). \textbf{(d)} Given samples at (1,2), the next observation with entropy [utility] $\geq$ the maximum entropy observation in $U_R$ minus $\lambda$ occurs at (3). }
\label{fig:period}
\end{figure}

To select the next observation in the stream to sample, the periodic secretary algorithm computes the utility of each observation in the reference set ${U}_R$ and finds the observation with the highest utility in the reference set given previously selected samples. Then, the algorithm samples the next observation in the data stream with utility greater than the maximum utility observation in the reference set, minus some constant threshold parameter $\lambda$ that accounts for noise in the periodic function. In a sense that we derive explicitly in Section \ref{sec:proof}, we can expect to see an observation of sufficient utility with high probability because our data are approximately periodic, and periodic observations produce periodic observation utilities. Given this new sample, the utility of observations in the reference set may have changed. We find the new maximum utility observation in the reference set conditioned on the new sample set, and select the next observation in the data stream within some $\lambda$ of this maximum. This procedure repeats until $k$ samples have been collected or the end of the data stream is reached. The procedure is formalized in Algorithm \ref{alg:alg1} and depicted visually in Figure \ref{fig:period}. We discuss the effect of the parameter $\lambda$ on the algorithm's performance in Section \ref{sec:exp}.


\section{Theoretical Algorithm Performance}
\label{sec:proof}

In this section, we analyze the performance of the periodic secretary algorithm as a function of the variables in our model: the utility noise $\sigma_u^2$, the number of periods in the data $\left\lfloor N/T \right \rfloor$, the number of samples selected $k$, and threshold parameter $\lambda$. We show that when the number of periods in the dataset is large compared to the number of samples selected, the gap between the performance of the periodic secretary algorithm and the optimal offline solution grows slowly with the length of the dataset as $O(\sqrt{\text{log}\left\lfloor N/T \right \rfloor})$. When the number of samples is much larger then the number of periods, our bound decreases quickly as $k$ grows, as $O(\left\lfloor N/T \right \rfloor / k )$. Although the algorithm will likely outperform this bound for specific utility functions, this is the tightest bound we could derive for general utility functions and is commensurate with bounds provided by e.g. the submodular secretary algorithm \cite{Bateni2010} when the number of periods divided by the number of samples is on the order of $(1 - 1/e) / 7$. These conclusions follow directly from Theorem 1, proven at the end of this section. However, we first provide the following three useful lemmas.

Let $\mathcal{A^*}$ be the optimal sample set according to Eq. (\ref{eq:utility}) and $\mathcal{A}$ be the set returned by the periodic secretary algorithm. We refer to the first $T$ observations in the stream as the reference set ${U}_R$. Let $\mathcal{A}_m \subseteq \mathcal{S}$ be the current set of $m$ observations sampled by the algorithm and $f_{\mathcal{A}_m}(\mathbf{x})$ be the marginal gain of adding observation $\mathbf{x}$ to set $\mathcal{A}_m$, i.e. $f(\mathcal{A}_m \cup \mathbf{x}) - f(\mathcal{A}_m)$.


\vspace{0.10cm}
\noindent
\textbf{Lemma 1.} In each iteration of the periodic secretary algorithm, the expected utility of the sample selected by the periodic secretary algorithm $\mathbf{x}_{s}^*$ from approximately periodic data of length $N$ with period $T$ and utility noise $\sigma_u^2$, given a previously selected sample set $\mathcal{A}_m$, is lower bounded by: 
\begin{equation}
\mathbb{E}[f_{\mathcal{A}_{m}}(\mathbf{x}_{s}^*)] \geq \mathbb{E}[f_{\mathcal{A}_{m}}(\mathbf{x}^*)] - \Big( \lambda + \sqrt{2\sigma_u^2 \text{ log}\left\lfloor \frac{N}{T} \right \rfloor} \Big),     
\end{equation}
where $\mathbf{x}^*$ is the point with globally maximum utility.

Lemma 1 bounds how suboptimal the sample selected by the periodic secretary algorithm can be compared to the globally optimal sample. The detailed proof of Lemma 1 is included in the Appendix.


\vspace{0.10cm}
\noindent
\textbf{Lemma 2.} A set $\mathcal{A}$ of $k$ samples chosen according to the periodic secretary algorithm will have utility:
\begin{equation}
\mathbb{E}[f(\mathcal{A})] \geq (1 - \frac{1}{e}) \left( f(\mathcal{A^*}) - k \cdot \Big( \lambda + \sqrt{2\sigma_u^2 \text{ log}\left\lfloor \frac{N}{T} \right \rfloor} \Big) \right).
\end{equation}

Lemma 2 states that a set of $k$ samples chosen with suboptimality bounded as in Lemma 1 also has bounded suboptimality. The detailed proof of Lemma 2 is included in the Appendix. Lemma 2 assumes that the periodic secretary algorithm succeeds in sampling $k$ times, as will be the case when the utility noise $\sigma_u^2$ is small and the length of the data stream is large. However, given a finite data stream of length $N$, it is possible to fail to select all $k$ samples.  


\vspace{0.10cm}
\noindent
\textbf{Lemma 3.} In an approximately periodic data stream with period $T$ and utility noise $\sigma_u^2$ of length $N$, the expected number of samples selected by the periodic secretary algorithm is:
\begin{align}
    \mathbb{E}[\# Success] \geq \text{min} \left(k, Q(-\lambda / \sigma_u^2) \left \lfloor {\frac{N}{T}}\right \rfloor  \right).
\label{eq:exp}
\end{align}

\vspace{0.10cm}
\noindent
\textbf{Proof of Lemma 3.} The probability of encountering an observation in period $n$ of the data which meets or exceeds the utility threshold for a given iteration of the periodic secretary algorithm and is therefore sampled is:
\begin{equation}
\begin{split}
\mathbf{Pr}(Success) \geq & \mathbf{Pr}\Big(f(\mathbf{x}_{i + nT}) \geq f(\mathbf{x}_{r}^*) - \lambda \Big) \\ 
\geq & Q(-\lambda / \sigma_u^2),
\end{split}
\end{equation}
where $Q(\cdot)$ is the standard Gaussian tail probability. In a data stream of length $N$, there are $\left \lfloor {\frac{N}{T}}\right \rfloor$ total periods and the expected number of successes is the number of periods multiplied by the probability of success in each period. 

\vspace{0.10cm}
\noindent
\textbf{Theorem 1.} Given a sample set $\mathcal{A}$ selected by the periodic secretary algorithm from a data stream of length $N$ that is approximately periodic with period $T$ and utility noise $\sigma_u^2$, the expected utility of $\cal A$ is less than the utility of the optimal set $\mathcal{A^*}$ by a factor which depends the number of samples selected $k$ and parameter $\lambda$:
\begin{equation}
\begin{split}
    \mathbb{E}[f(\mathcal{A})] \geq &\frac{\text{min}(k \text{, }Q(-\lambda / \sigma_u^2) \left \lfloor {\frac{N}{T}}\right \rfloor)}{k} \cdot \left(1 - \frac{1}{e}\right) \\
    & \left( f(\mathcal{A^*}) - k \cdot \left( \lambda + \sqrt{2\sigma_u^2 \text{ log}\left\lfloor \frac{N}{T} \right\rfloor} \right) \right),
\end{split}
\label{eq:t1}
\end{equation}
where $Q(\cdot)$ is the standard Gaussian tail probability. 

\vspace{0.10cm}
\noindent
\textbf{Proof.}  In Lemma 2, we showed that a set of $k$ samples selected using the periodic secretary algorithm has bounded suboptimality. In practice, for finite data streams, it is possible successfully sample less than $k$ times. Lemma 3 derives the expected number of samples the periodic secretary algorithm will accept in a data stream of length $N$. Combining Lemma 2 and 3 with the observation that for a monotone submodular function, the value of the first $a$ samples of $\mathcal{A}$ have utility of at least $\left \lfloor {\frac{a}{k}}\right \rfloor f(\mathcal{A})$, $a \leq k$, the expected utility of set $\cal A$ is given by Theorem 1.\hfill\(\qedsymbol\)


\section{EXPERIMENTS} 
\label{sec:exp}
\subsection{Using simulation to tune parameter $\lambda$}
\begin{figure}[t]
    \centering
    \includegraphics[width=7.8cm]{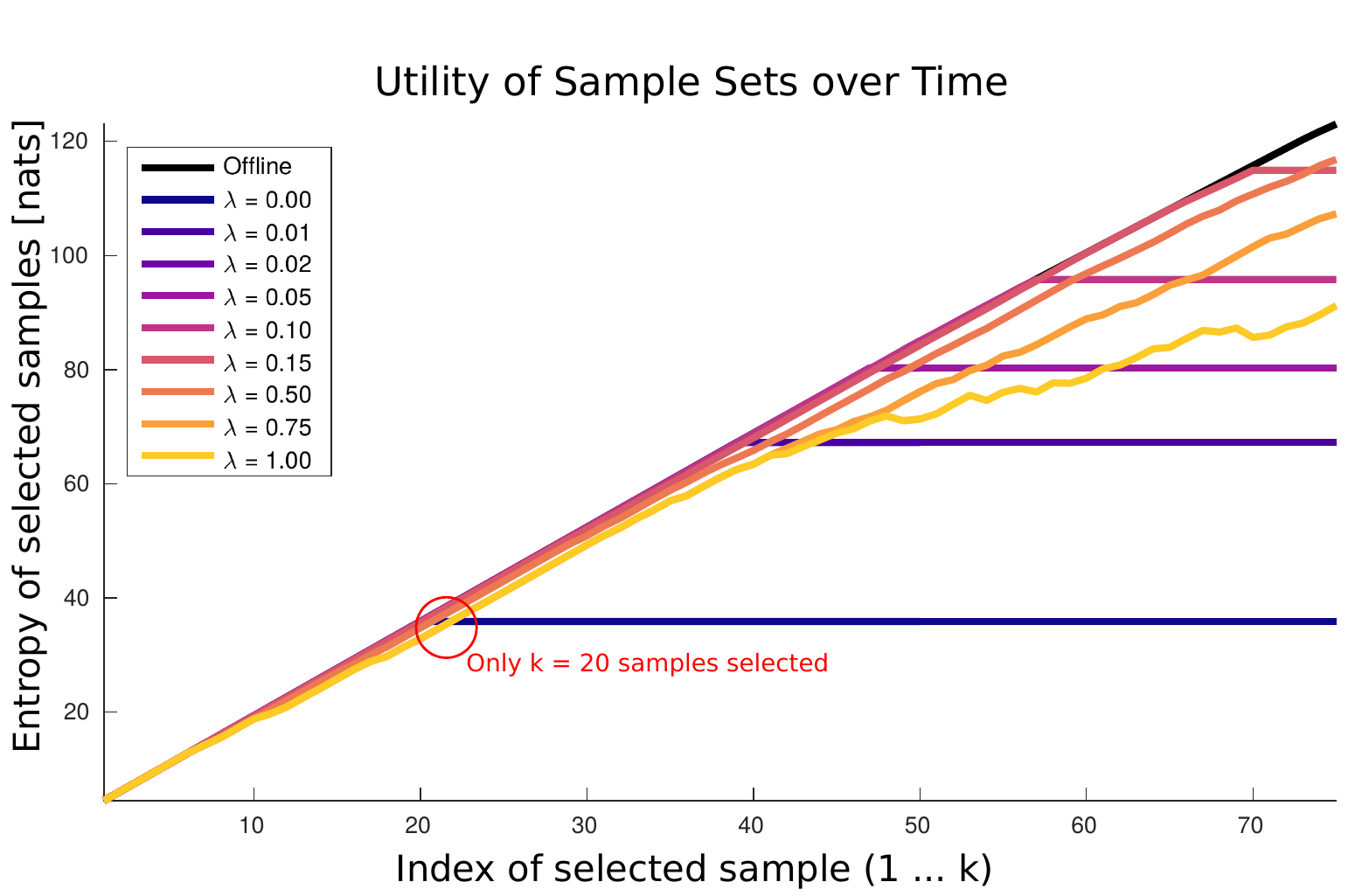}
    \caption{\textbf{Tuning parameter $\lambda$: } The utility of samples sets selected using the periodic secretary algorithm on the data stream $x(t) = \text{sin}(2 \pi t) + \text{sin}(3 \pi t)$ and periodic noise $\sigma_d^2 = 0.35$ for nine different values of $\lambda$, $\left\lfloor {N}/{T} \right\rfloor = 10$, and $k = 75$ with the entropy criterion. For small $\lambda$, the algorithm chooses high utility samples, but is unable to successfully sample $k$ times. For medium $\lambda$, the algorithm selects $k$ samples with utility very near that of the offline algorithm. For large $\lambda$ the algorithm successfully samples $k$ times, but the samples are of low utility. For this dataset, $\lambda$ should be set to 0.50 for best performance.} 
\label{fig:sim}
\end{figure}

The submodular secretary algorithm has one tunable parameter $\lambda$ that mediates the trade-off between selecting more, lower quality samples and selecting fewer, higher quality samples in a noisy data stream. Generally, for large $\lambda$, the expected number of samples selected will grow to $k$, but the utility of the selected samples will decrease. Smaller $\lambda$ will cause the samples in $\mathcal{A}$ to be closer to their optimal utility values, but the algorithm may fail to select all $k$ samples in a noisy, short data stream. Generally, $\lambda$ should be tuned to maximize Eq. (\ref{eq:t1}) based on the noise parameters of the periodic phenomena and the length of the data stream. We believe that it may be possible to do this maximization in closed form, but leave this as an open question for future work. It is also possible to tune $\lambda$ empirically by simulating data drawn from the periodic phenomena using the known period and periodic noise values and then selecting the $\lambda$ which produces the largest average utility across these simulated data streams. We demonstrate this process using samples drawn from an arbitrary approximately periodic function for nine different values of $\lambda$ in Figure \ref{fig:sim}. 

\subsection{MVCO Experiments}
\begin{figure}[b]
    \centering
    \includegraphics[width=8.0cm]{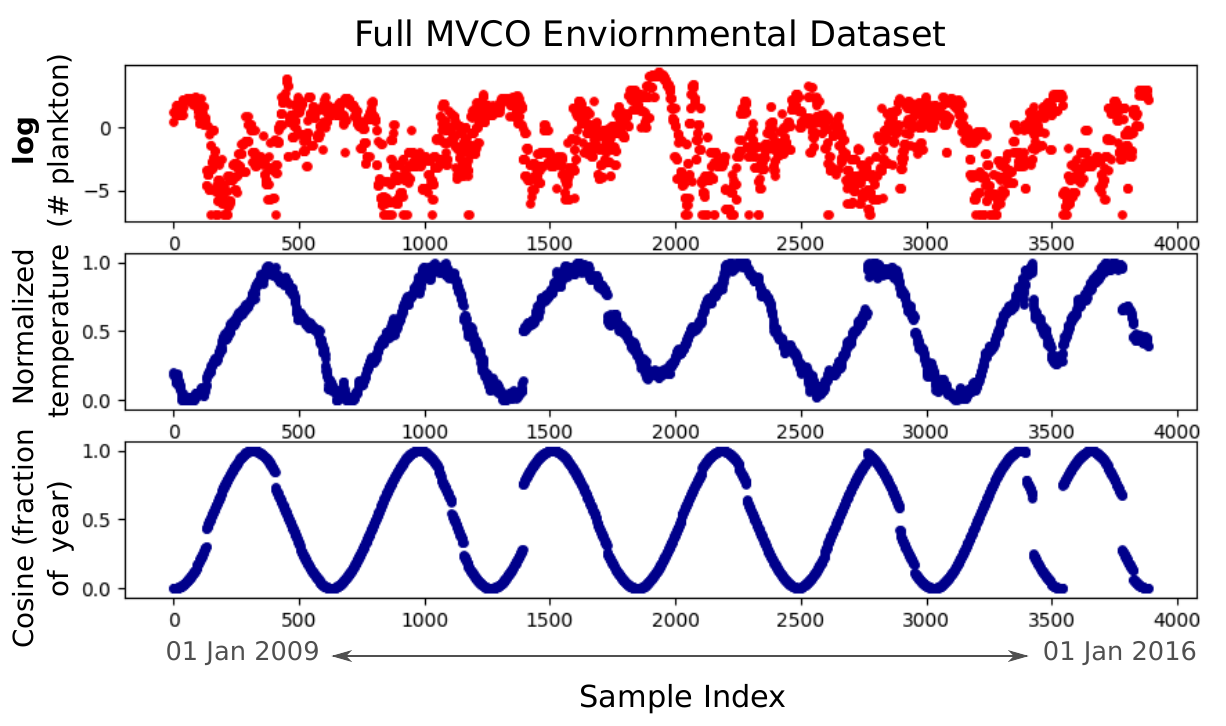}
    \caption{\textbf{MVCO Environmental Dataset:} The environmental dataset collected on the Martha's Vineyard Coastal Observatory from Jan 2009 to Jan 2016, averaged over half-day segments. The platform was equipped with the IFCB device \cite{Olson2007}, which allowed ground truth \textit{Guinardia flaccida} concentrations to be measured (red). Only the environmental data (blue) are available to the sample selection algorithms.}
    \label{fig:raw}
\end{figure}

We apply the periodic secretary algorithm with the entropy utility function to select water samples from a stream of potential samples observed by a marine sensor on the Martha's Vineyard Coastal Observatory from January 2009 to January 2016 \cite{Olson2007}. This stationary marine sensor is equipped with $k$ single-use water samplers. The scientific objective is to collect water samples that give the best understanding of the seasonal dynamics of the plankton species \textit{Guinardia flaccida}. The prevalence of this plankton species is known to vary with time of year (it is a winter blooming plankton) and water temperature (during warm winters, the species tends to be more numerous than during cold winters). However, the sensor is unable to measure the plankton present in the water stream in real-time.  Instead, the sensor can measure the temperature of the surrounding water and the day of year, and must decide to collect a sample based on these environmental covariates. In this stationary setting, the sensor is not choosing sample locations in geographic space but instead in the space of its environmental sensors. Throughout its deployment, the sensor will observe a stream of points in this environmental space, and must choose to take water samples in the environmental conditions which are the most informative about the plankton species of interest. This seven-year dataset and ground truth \textit{Guinardia flaccida} counts (unknown to the algorithm) are shown in Figure \ref{fig:raw}. 

\begin{figure}[t]
    \centering
    \includegraphics[width=8.0cm]{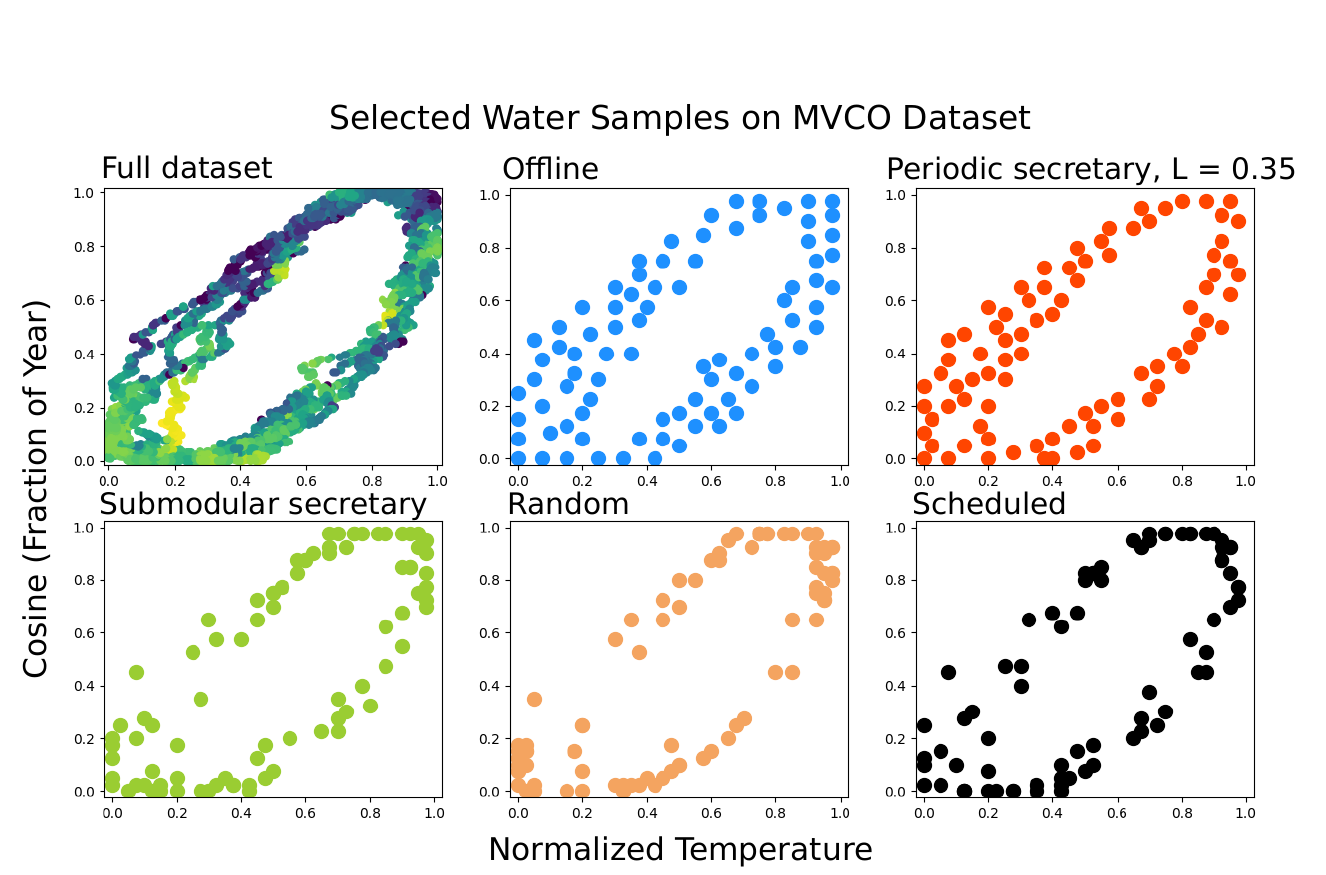}
  \caption{\textbf{Samples selected from the MVCO dataset:}   
  The \{temperature, $\cos$(fraction of year)\} samples selected from the full dataset (upper left, colored by the ground-truth plankton counts). The periodic secretary algorithm chooses samples which provide the most dense coverage of the environmental observation space. The quality of predictions in unknown environmental conditions will depend on having sampled a nearby point in \{temperature, $\cos$(fraction of year)\} space. Gaps in the sample coverage, such as those seen in the bottom three plots, will cause large uncertainty and poor predictions of plankton counts in those regions.} 
    \label{fig:samps}
\end{figure}

Given that these environmental data are known to be periodic on an annual basis, we apply the periodic secretary algorithm to select 84 samples (equivalent to 12 samples per year for seven years using a scheduled sampler) from the data stream using the entropy criterion with a GP model.  We also select sample locations using the submodular secretary algorithm \cite{Bateni2010}, a scheduled sampling algorithm commonly used in practical sensing deployments (sampling every $\frac{N}{k}$ samples), and random sampling as baselines.  We use the offline greedy algorithm \cite{Nemhauser1978} to provide an upper bound. 

\subsection{MVCO Sampling Results}
The selected samples for each approach are shown in Figure \ref{fig:samps} along with the complete dataset colored by the ground-truth plankton counts. The periodic secretary algorithm selects samples which provide the most dense coverage of the observation space. The quality of plankton count predictions in unknown environmental conditions will depend on having sampled a nearby point in \{temperature, $\cos$(fraction of year)\} space. Large gaps in the sampled locations will cause lower entropy reduction and poorer predictions at those locations; these gaps are evident in the submodular, scheduled and random sampling strategies. 


\begin{figure}[]
    \centering
    \includegraphics[width=7.2cm]{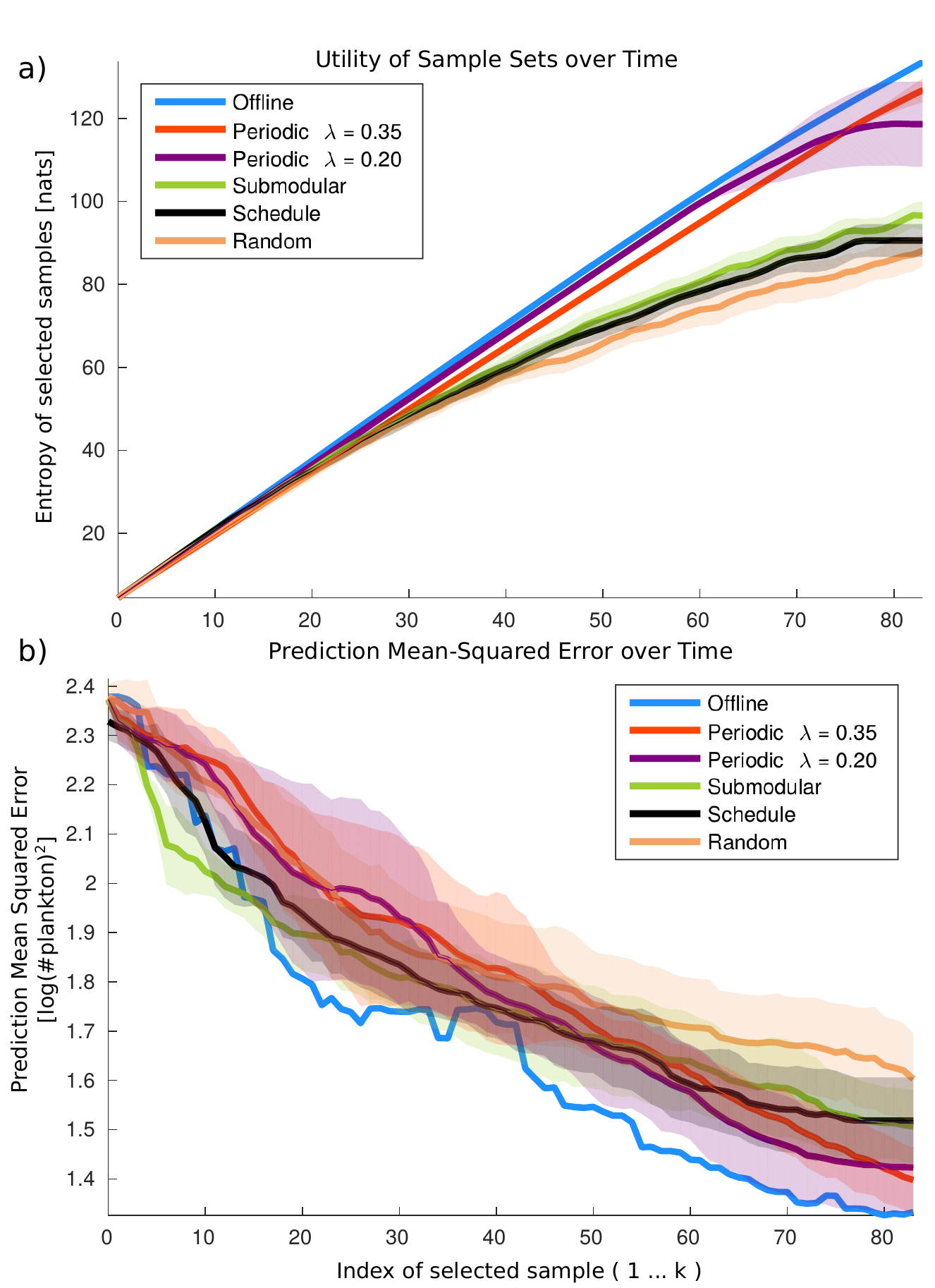}
   \caption{\textbf{Quantitative results on the MVCO dataset:} The mean value and standard deviation across the 50 runs of the periodic secretary algorithm on random permutations of the yearly data in the MVCO dataset. (a) The entropy reduction achieved as each of the $k = 84$ total samples are selected. The periodic secretary algorithm achieves the highest entropy reduction among the streaming algorithms. However, the algorithm with poorly tuned $\lambda$ reaches the end of the stream without selecting all samples. The periodic secretary algorithm with well-tuned $\lambda$ stays very close to the upper bound set by the greedy algorithm. (b) Using the selected samples, plankton counts at unknown locations are predicted on a held-out test set. The prediction mean-squared error decrease as samples are selected and is minimized using the periodic secretary algorithm with well tuned $\lambda$.} 
    \label{fig:quant}
\end{figure}

To quantify this result, we compare the entropy reduction achieved by samples selected using the periodic secretary algorithm to samples selected by the baselines and the offline upper bound (the entropy reduction should be maximized). The mean utility and one standard deviation values for each algorithm are shown in Figure \ref{fig:quant}(a) for 50 random permutations of the yearly data in the MVCO dataset. For small sample sets, all six algorithms produce similar results. After selecting around 30 samples, the periodic secretary algorithms begin to surpass the other streaming algorithms. The submodular secretary algorithm, which represents the current state-of-the-art in streaming, irrevocable sample selection for information-theoretic utility functions, never does much better than a scheduled algorithm. After selecting $70$ samples, the periodic secretary algorithm with poorly tuned $\lambda$ reaches the end of the stream without selecting all $k = 84$ samples. The periodic secretary algorithm with well-tuned $\lambda$ stays close to the upper bound set by the offline algorithm.


Figure \ref{fig:quant}(a) demonstrates that samples selected by the periodic secretary algorithm achieve the highest entropy reduction across the environmental observation space. Intuitively, this means that we can use these samples to do inference about plankton concentrations in unknown environmental conditions. To test this assumption, we quantify how well the representative samples selected by each algorithm can be used to predict \textit{Guinardia flaccida} concentrations on a held out test set of \{temperature, $\cos$(fraction of year)\} environmental conditions (the prediction mean-squared error should be minimized). Figure \ref{fig:quant}(b) shows that on average the sample sets selected by the periodic secretary algorithm produce more accurate predictions of plankton counts then all other streaming algorithms. Note that choosing points according to the entropy criterion is a nearly optimal strategy from an information theoretic perspective when trying to reduce prediction error, but higher entropy reduction will not necessary directly equate to lower mean-squared prediction error for a specific dataset. This is why there are places in Figure \ref{fig:quant}(b) where an algorithm with lower entropy reduction achieves lower prediction mean-squared error.  


\section{DISCUSSION AND CONCLUSION}
This paper presents a novel algorithm for online, irrevocable sample selection from periodic phenomena. We prove that the periodic secretary algorithm selects sample sets according to any monotone submodular set function with bounded suboptimality. For short data streams, where the number of periods is small compared to the number of samples to be selected, the performance of the periodic secretary algorithm depends on the choice of utility function and setting the parameter $\lambda$ appropriately. However, we demonstrate that for the entropy criterion, this dependence is only evident when the number of samples is much larger then the number of periods. and provide methods to tune $\lambda$. 



The periodic secretary algorithm is a robust and versatile tool that can be applied in a variety of real-time applications; many real-world periodic data streams can be considered approximately periodic, so long as period-to-period variation can be modeled as Gaussian noise with some covariance $\Sigma_d$. The algorithm is also robust to noisy estimates of the period length $T$, requiring only that the algorithm's reference set includes one complete period of the data. Our work extends previous results in information theoretic sample selection and adapts classical secretary algorithms to data domains that produce  periodic spatially and/or temporally correlated data streams, such as robotics and environmental monitoring.  Although we focus on periodic phenomena, we believe techniques similar to those presented here could be used to provide performance bounds for irrevocable sample selection from data streams with other types of spatiotemporal structure, and hope that this work will serve as a foundation for developing secretary algorithms that can be applied to these intersting data domains.

\bibliographystyle{IEEEtran}
\bibliography{icra_genevieve,girdhar}

\section{APPENDIX}
We include some technical proofs removed from the body of the paper in the interest of space:

\vspace{0.10cm}
\noindent
\textbf{Proof of Lemma 1.} Given that the maximum utility observation in the reference set $\mathbf{x}_r^*$ occurs at index $i$, the expected difference in utility between  $\mathbf{x}_r^*$ and the the maximum utility sample in the entire data stream $\mathbf{x^*}$ will be maximized when $\mathbf{x^*}$ occurs at index $i + nT$ for some $n$, $0 \geq n \geq \left\lfloor \frac{N}{T} \right \rfloor$. Because our data are approximately periodic, we know that $f_{\mathcal{A}_{m}}(\mathbf{x}_{i + nT}) \sim \mathcal{N}(f_{\mathcal{A}_{m}}(\mathbf{x}_r^*), \sigma_u^2)$ for $n = \{0, \dots, \frac{N}{T}\}$ and the global maximum $\mathbf{x}^* = \text{max } \{\mathbf{x}_{i+nT} \mid n = 0, \dots, \left\lfloor \frac{N}{T} \right\rfloor\}$ i.e. the maximum of $n$ i.i.d. draws from a normal distribution with mean $f_{\mathcal{A}_{m}}(\mathbf{x}_r^*)$ and variance $\sigma_u^2$. Therefore, the expected difference between$f_{\mathcal{A}_{m}}(\mathbf{x}^*)$ and $f_{\mathcal{A}_{m}}(\mathbf{x}_r^*)$ is no larger than the expectation of the maximum of $n$ samples drawn from a mean-zero Gaussian \cite{Kamath}:
\begin{equation}
\mathbb{E}[f_{\mathcal{A}_{m}}(\mathbf{x^*}) - f_{\mathcal{A}_{m}}(\mathbf{x}_r^*)] \leq \sqrt{2\sigma_u^2 \text{ log} \left\lfloor \frac{N}{T} \right \rfloor}.
\end{equation}

The sample that the algorithm selects $\mathbf{x}_s^*$ will have utility $f_{\mathcal{A}_{m}}(\mathbf{x}_s^*) = f_{\mathcal{A}_{m}}(\mathbf{x}_r^*) - \lambda$. 

\vspace{0.10cm}
\noindent
\textbf{Proof of Lemma 2.} Following the general proof in \cite{Horel2016}:
\begin{align} 
    \label{eq:line1}
    f(\mathcal{A}^*)  & \leq  f(\mathcal{A}_{m-1}) + \sum_{\mathbf{x} \in \mathcal{A^*} \setminus \mathcal{A}_{m-1}} f_{\mathcal{A}_{m-1}}(\mathbf{x})\\
    \label{eq:line2}
    & \leq f(\mathcal{A}_{m-1}) + \sum_{\mathbf{x} \in \mathcal{A^*} \setminus \mathcal{A}_{m-1}} f(\mathcal{A}_{m}) - f(\mathcal{A}_{m-1}) +  c\\
    \label{eq:line3}
    & \leq f(\mathcal{A}_{m-1}) + k \cdot (f(\mathcal{A}_{m}) - f(\mathcal{A}_{m-1}) + c),
\end{align}
where $ c = \lambda + \sqrt{2\sigma_u^2 \text{ log}\left\lfloor \frac{N}{T} \right \rfloor}$. The first line (\ref{eq:line1}) follows directly from $f(\cdot)$ being a monotone submodular set function \cite{Horel2016}, the second (\ref{eq:line2}) from Lemma 1, and the third (\ref{eq:line3}) because $|\mathcal{A}^*| \leq k$.  Subtracting $k \cdot f(\mathcal{A}^*)$ from both sides:
\begin{align}
    f(\mathcal{A}_m) - f(\mathcal{A^*}) & \geq \frac{k - 1}{k} (f(\mathcal{A}_{m-1}) - f(\mathcal{A^*})) - c,
\end{align}
which implies by induction, with $f(\emptyset) = 0$:
\begin{align}
    f(\mathcal{A}_i) & \geq \left( 1 - (1 - \frac{1}{k})^i \right) \Big(f(\mathcal{A^*}) - k \cdot c \Big).
\end{align}
Lemma 2 is achieved by setting $i = k$, and using the identity $(1 - \frac{1}{k})^k \leq \frac{1}{e}$.

\end{document}